\documentclass[12pt]{article}
\usepackage{amsfonts}
\usepackage{amsmath, amsthm}
\usepackage{epsfig}
\usepackage{rotating}
\usepackage{amsmath,empheq}
\usepackage{authblk}
\usepackage{mathtools}

\usepackage[toc,page]{appendix}

\usepackage{algorithm,algorithmic}
\usepackage{bm}

\usepackage{authblk}

\usepackage{amsmath,verbatim,color,amssymb,epsfig}
\usepackage{bm}
\usepackage{amsfonts}
\usepackage{subfig}
\usepackage{epsfig}
\usepackage{multirow}
\usepackage{graphicx}
\usepackage{array}
\usepackage[table]{xcolor}
\definecolor{Gray}{gray}{0.80}
\usepackage{lscape}
\usepackage{cite}
\usepackage{natbib}
\usepackage[normalem]{ulem}
\usepackage[colorlinks=true,urlcolor=blue,citecolor=purple,linkcolor=blue,bookmarks=true]{hyperref}
\DeclareMathOperator*{\argmin}{argmin} 
\DeclareMathOperator*{\median}{median}
\DeclareMathOperator*{\sign}{sign}
\DeclareMathOperator*{\abs}{abs}

\usepackage{caption}
\begin{document}
\def\eqx"#1"{{\label{#1}}}
\def\eqn"#1"{{\ref{#1}}}
\newtheorem{remark}{Remark}
\makeatletter 
\@addtoreset{equation}{section}
\makeatother  

\def\yincomment#1{\vskip 2mm\boxit{\vskip 2mm{\color{red}\bf#1} {\color{blue}\bf --Yin\vskip 2mm}}\vskip 2mm}
\def\squarebox#1{\hbox to #1{\hfill\vbox to #1{\vfill}}}
\def\boxit#1{\vbox{\hrule\hbox{\vrule\kern6pt
          \vbox{\kern6pt#1\kern6pt}\kern6pt\vrule}\hrule}}

\def\theequation{\thesection.\arabic{equation}}
\newcommand{\ds}{\displaystyle}

\newcommand{\bJ}{\mbox{\bf J}}
\newcommand{\bF}{\mbox{\bf F}}
\newcommand{\bM}{\mbox{\bf M}}
\newcommand{\bR}{\mbox{\bf R}}
\newcommand{\bZ}{\mboxZ}
\newcommand{\bX}{\mbox{\bf X}}
\newcommand{\bx}{\mbox{\bf x}}
\newcommand{\bQ}{\mbox{\bf Q}}
\newcommand{\bH}{\mbox{\bf H}}
\newcommand{\bh}{\mbox{\bf h}}
\newcommand{\bz}{\mboxZ}
\newcommand{\ba}{\mbox{\bf a}}
\newcommand{\be}{\mbox{\bf e}}
\newcommand{\bG}{\mboxG}
\newcommand{\bB}{\mbox{\bf B}}
\newcommand{\bb}{\mbox{\bf b}}
\newcommand{\bA}{\mbox{\bf A}}
\newcommand{\bC}{\mbox{\bf C}}
\newcommand{\bI}{\mbox{\bf I}}
\newcommand{\bD}{\mbox{\bf D}}
\newcommand{\bU}{\mbox{\bf U}}
\newcommand{\bc}{\mbox{\bf c}}
\newcommand{\bd}{\mbox{\bf d}}
\newcommand{\bs}{\mbox{\bf s}}
\newcommand{\bS}{\mbox{\bf S}}
\newcommand{\bV}{\mbox{\bf V}}
\newcommand{\bv}{\mbox{\bf v}}
\newcommand{\bW}{\mbox{\bf W}}
\newcommand{\bw}{\mbox{\bf w}}
\newcommand{\bg}{\mboxG}
\newcommand{\bu}{\mbox{\bf u}}
\def\bb{{\bf b}}

\newcommand{\bcU}{\boldsymbol{\cal U}}
\newcommand{\bbeta}{\boldsymbol{\beta}}
\newcommand{\bdelta}{\boldsymbol{\delta}}
\newcommand{\bDelta}{\boldsymbol{\Delta}}
\newcommand{\boldeta}{\boldsymbol{\eta}}
\newcommand{\bxi}{\boldsymbol{\xi}}
\newcommand{\bGamma}{\boldsymbol{\Gamma}}
\newcommand{\bSigma}{\boldsymbol{\Sigma}}
\newcommand{\balpha}{\boldsymbol{\alpha}}
\newcommand{\bOmega}{\boldsymbol{ R}}
\newcommand{\btheta}{\boldsymbol{\theta}}
\newcommand{\bmu}{\boldsymbol{\mu}}
\newcommand{\bnu}{\boldsymbol{\nu}}
\newcommand{\bgamma}{\boldsymbol{\gamma}}

\newtheorem{thm}{Theorem}[section]
\newtheorem{lem}{Lemma}[section]
\newtheorem{rem}{Remark}[section]
\newtheorem{cor}{Corollary}[section]
\newcolumntype{L}[1]{>{\raggedright\let\newline\\\arraybackslash\hspace{0pt}}m{#1}}
\newcolumntype{C}[1]{>{\centering\let\newline\\\arraybackslash\hspace{0pt}}m{#1}}
\newcolumntype{R}[1]{>{\raggedleft\let\newline\\\arraybackslash\hspace{0pt}}m{#1}}

\newcommand{\tabincell}[2]{\begin{tabular}{@{}#1@{}}#2\end{tabular}}
\def\correspondingauthor{\footnote{zhe13@fordham.edu}}

\title{\bf Gradient Boosting Machine: A Survey}

\author[1,2]{Zhiyuan He\correspondingauthor{}}
\author[1, 3]{Danchen Lin}
\author[1]{Thomas Lau}
\author[1, 4]{Mike Wu}
\affil[1]{Point Zero One Technology}
\affil[2]{Fordham University}
\affil[3]{Imperial College London}
\affil[4]{Massachusetts Institute of Technology}

\maketitle
\begin{abstract}
	In this survey, we discuss several different types of gradient boosting algorithms and illustrate their mathematical frameworks in detail: \emph{1.} introduction of gradient boosting leads to \emph{2.} objective function optimization, \emph{3.} loss function estimations, and \emph{4.} model constructions. \emph{5.} application of boosting in ranking.
\vspace{0.5cm}
\end{abstract}

\section{Introduction}

Proposed by \cite{Freund1997}, boosting is a general issue of constructing an extremely accurate prediction with numerous roughly accurate predictions. Addressed by \cite{Friedman2001,Friedman2002} and \cite{Natekin2013}, the Gradient Boosting Machines (GBM) seeks to build predictive models through back-fittings and non-parametric regressions. Instead of building a single model, the GBM starts by generating an initial model and constantly fits new models through loss function minimization to produce the most precise model \citep{Natekin2013}.

This survey concentrates on the mathematical derivations of the gradient boosting algorithms. In Section 2, we analyze the optimization methods for parametric and non-parametric models. Section 3 covers the definitions of different types of loss functions. In Section 4, we present different types of boosting algorithms, while in Section 5, we explore the combination of boosting algorithms and ranking algorithms to rank the real-world data. 

\section{Basic Framework}
The ultimate goal of the GBM is to find a function $F(x)$, which minimize its loss function $L(y,F(x))$ as
\begin{equation*}\label{MinimizingLossFunction}
F^*= \argmin_{F} E_{y,x}L(y,F(x)),
\end{equation*}
through iterative back-fitting.

\subsection{Numerical Optimization}
By definition, a boosted model is a weighted linear combination of the base learners
\begin{equation*}\label{AdditiveF}
F(x;\{\beta_{m},a_{m}\}_{1}^{M}) = \sum_{m=1}^{M} \beta_{m}h(x:a_{m}),
\end{equation*}
where $h(x;a)$ is a base learner parameterized by $a$. Regarded as weak learners, the base learners produce hypotheses that only predict slightly better than random guessing, and it was proved that recursive learning with weak learners can perform just as good as a strong learning algorithm \citep{Schapire1990}. 

If the base learner is a regression tree, the parameter $a$ is usually the splitting nodes of tree branches \citep{Friedman2002}. Tree-based models divide the input variable space into regions and apply a series of rules to identify the regions that have the strongest responses to the inputs \citep{Elith2008}. Each region is then fitted with a regression tree taking the mean response of the observations \citep{Elith2008}. Decision trees are constructed through binary splits, and recursive splits generate a large tree, which is then pruned to drop out the weak branches.

The optimization process can be written as follows
\begin{equation*}\label{P^*}
P^* = \argmin_{p} \Phi(P)
\Phi(P) = E_{y,x}L(y,F(x;P))
F^*(x) = F(x;P^*),
\end{equation*}
where $P^* = \sum_{m=0}^{M} p_{m},$
and $p_{m}$ are the consecutive boosting steps. 

One of the approaches to generate these steps, $p_{m}$, is to use the steepest-descent algorithm by calculating the gradient
\begin{equation*}\label{Gradient}
g_{m} = \{g_{jm}\} = \left\{ \frac{\delta \Phi(P)}{\delta P_{j}}\bigg|_{P=P_{m-1}}\right\},
\end{equation*}
where $P_{m-1} = \sum_{i=0}^{m-1} p_{i}.$ 
The boosting step in the previous function is given by
\begin{equation*}\label{BoostingStep}
p_{m} = -\rho_m g_{m},
\end{equation*}
and $\rho_{m} = \argmin_{\rho} \Phi(P_{m-1} - \rho g_{m})$ is the line search on the direction of steepest-descent. 

In the non-parametric case, the function $F$ is solved by minimizing
\begin{equation*}\label{NonParametric}
\Phi(F) = E_{y,x}L(y,F(x)) = E_{x}[E_{y}(L(y,F(x))) | x] = E_{y}[L(y,F(x)) | x],
\end{equation*}
and the optimum is reached at
$F^*(x) = \sum_{m=0}^{M} f_{m}(x),$ 
where $f_{m}(x) = -\rho_{m}g_{m}(x).$

The gradient of the non-parametric model is
\begin{equation*}\label{NonParametric Gradient}
g_{m}(x) = \frac{\delta \Phi(F(x))}{\delta F(x)}\bigg|_{F(x) =F_{m-1}(x)} =  \frac{\delta E_{y} [L(y,F(x)) | x ]}{\delta F(x)}\bigg|_{F(x) = F_{m-1}(x)}.
\end{equation*}
The differentiation and integration of the gradient function can be switched interchangeably, and the gradient function can be simplified to
\begin{equation*}\label{Simplified Gradient}
g_{m}(x) = E_{y} \left[\frac{\delta \Phi(F(x))}{\delta F(x)}\bigg|x\right]_{F(x) =F_{m-1}(x)}. 
\end{equation*}
According to the above equation, the line search is solved as
\begin{equation*}\label{NonParametric LineSearch}
\rho_{m} = \argmin_{\rho} E_{y,x}L(y,F_{m-1}(x) - \rho g_{m}(x)).
\end{equation*}

\subsection{Finite Dataset}
With finite number of samples, a non-parametric function $F(x)$ can be obtained by a greedy stage-wise algorithm.
Different from stepwise strategy, the stage-wise strategy does not make adjustments to the previous boosting steps, and thus can be demonstrated as
\begin{equation*}\label{StageWise Parametric}
(\beta_{m},a_{m}) = \argmin_{\beta,a} \sum_{i=1}^{N} L(y_{i}, F_{m-1}(x_{i}) + \beta h(x_{i};a)).
\end{equation*}
Thus, $F(x)$ can be obtained iteratively
\begin{equation*}
F_{m}(x) = F_{m-1}(x) + \beta_{m} h(x;a_{m}).
\end{equation*}

\section{Estimation}
Under the framework of GBM, different loss functions can be applied to solve different tasks \citep{Natekin2013, Koenker2001,Friedman2002}. 

\subsection{Continuous Response} 
$L_1$ loss function, known as Laplacian loss, is presented as
\begin{equation*}\label{Laplacian}
L(y,F)_{L_{1}} = |y-F|,
\end{equation*}
which is the absolute value of residuals between the explained variable $y$ and the predictive function $F$, while the most commonly used squared-error $L_2$ loss function is defined as
\begin{equation*}\label{Squared-errorLoss}
L(y,F)_{L_{2}} = \frac{1}{2}(y-F)^2.
\end{equation*}

In addition, the Huber loss function that merges $L_1$ and $L_2$ loss functions described above can be a robust alternative to the $L_1$ loss function
\begin{equation*}\label{Huber}
L(y,F)_{Huber,\delta} = \left\{\begin{array}{rcl}\frac{1}{2}(y-F)^2 \ \ \ |y-F| \leq \delta\\ \delta(|y-F|-\frac{\delta}{2}) \ \ \ |y-F| > \delta\end{array}\right..
\end{equation*}

Quantile loss is inevitably handy in the situations of ordering and sorting because of its robustness, which is framed as
\begin{equation*}\label{Quantile}
L(y,f)_{\alpha} = \left\{\begin{array}{rcl} (1-\alpha)|y-f| \ \ \ y-f \leq 0\\\alpha|y-f| \ \ \ y-f > 0\end{array}\right.,
\end{equation*}
where $\alpha$ designates the targeted quantile in the conditional distribution. The loss function can be degenerated into the $L_1$ loss by taking $\alpha = 0.5$. 

\subsection{Categorical Response}
There are two loss functions designed for categorical response that are widely used, namely the Bernoulli loss function and the exponential loss function. The Bernoulli loss function is formulated as
\begin{equation*}\label{Bernoulli}
L(y,F)_{Bern} = \log(1+\exp(-2\overline{y}F)),
\end{equation*}
while in Adaboost algorithm, the same alteration on $y$ variable in Bernoulli loss is applied on $y$ variable in exponential loss \citep{Natekin2013}
\begin{equation*}\label{Exponential}
L(y,F)_{Ada} = \exp(-\overline{y}F).
\end{equation*}

\section{Methodology}

\subsection*{AdaBoost}
Gradient boosting is a generalization of Adaboost. The design of Adaboost \citep{Freund1997}, the original boosting algorithm, is to find a hypothesis with low prediction error relative to a given distribution over the training samples. \cite{Freund1997} demonstrated their algorithm through a horse gambling example, where a gambler wishes to bet on the horse that has the greatest chance to win. In order to increase the winning probability of a bet, the gambler is encouraged to gather the expert opinions before placing a bet. Such a process of collecting information from different experts is similar to the ensemble of a class of poor classifiers. In Adaboost, each expert's opinion corresponds to a training set \citep{Wang2012}. Each sample is initialized with a weight, and the weights of the training sets are adjusted after each iteration, such that the weights of misclassified samples are increased, while the weights of correctly classified samples are decreased. 

In each iteration of boosting, the current weak learner of Adaboost chooses a weak hypothesis from the entire set of weak hypotheses instead of just the weak hypotheses that are currently found to the point. Since the search of an entire space of hypotheses can be enormous amount of work, it is often suitable to apply weak learners that approximately cover the whole set \citep{Collins2002}.

Boosting algorithms with certain modifications perform well under high bias and high variance settings. When weighted sampling is implemented for the training data, the performance of boosting is determined by its ability to reduce variance \citep{Friedman2000}. Meanwhile, boosting performance depends on bias reduction when the weighted sampling is replaced with weighted tree fitting \citep{Friedman2000}. 

Additionally, Adaboost is prone to cause model overfitting because of the exponential loss. The overfitting may be mitigated by minimizing the normalized sigmoid cost function in exchange \citep{Mason2000},
\begin{equation*}\label{Normalized Sigmoid}
C(F) = \frac{1}{m} \sum_{i=1}^{m} 1 - \tanh(\lambda y_{i} F(x_{i})).
\end{equation*}
In the above function, $F$ is a convex combination of weak hypotheses, and the parameter $\lambda$ measures the steepness of the margin cost function $c(z) = 1 - \tanh(\lambda z)$. Through their experiments, \cite{Mason2000} showed that a new boosting algorithm optimizing normalized sigmoid cost, called DOOM II, overall performed better than Adaboost. According to \cite{Mason1999}, AnyBoost is a general boosting algorithm that optimize gradient descent in an inner product space. The inner product space $S$, which is inclusive of all linear combinations of weak hypotheses, contains the weak hypotheses and their combination $F$. The inner product can thus be represented as, 
\begin{equation*}\label{Inner Product}
\langle F,G \rangle \coloneqq \frac{1}{m} \sum_{i=1}^{m} F(x_{i})G(x_{i}),
\end{equation*}
where $F$ and $G$ are the combinations of weak hypotheses that belong to the set of all linear combinations of weak hypotheses. Only AnyBoost algorithm that implies the inner product function and normalized sigmoid cost function is referred to DOOM II \citep{Mason2000}.

\subsection*{Arc-x4}
Arcing, a concept introduced by \cite{Breiman1996} and utilized in Adaboost, is a technique to adaptively reweighting the training samples. Arc-x4 \citep{Breiman1997} performs similarly to the original boosting in training error and generalization error reduction. At each boosting step, a new training sample is generated from the training set with probability 
\begin{equation*}\label{Arcx4 Probabilities}
p(n) = \frac{(1+m(n)^4)}{\sum(1+m(n)^4)},
\end{equation*}
where $m(n)$ is the number of misclassified cases. 

\subsection*{Least Squares Boost}
The least-squared loss function in continuous response is one of the most commonly used loss function. In parametrized model, the optimization using the least-squared loss has an equation
\begin{equation*}\label{LS Optimization}
(\rho_{m},a_{m}) = \argmin_{a, \rho} \sum_{i=1}^{N} [\widetilde{y_{i}} - \rho h(x_{i};a)]^2.
\end{equation*}
Solving for $F(x)$, we obtain a stage-wise model
\begin{equation*}\label{LS_Solution}
F_{m}(x) = F_{m-1}(x) + \rho_{m}h(x;a_{m}).
\end{equation*}

\subsection*{Logitboost}
Another well-known boosting Algorithm is Logitboost. Similar to other boosting algorithms, Logitboost adopts regression trees as the weak leaners. Deriving from the logistic regression, Logitboost takes the negative of the loglikelihood of class probabilities \citep{Li2012}. Defined as $p$, class probability is formulated as
\begin{equation*}\label{ClassProbability}
p_{i,k} = Pr(y_{i} = k | x_{i}) = \frac{e^{F_{i,k}(x_{i})}}{\sum_{s = 0}^{K-1} e^{F_{i,s}(x_{i})}},
\end{equation*}
where $y_{i}$ is the output vector and $X_{i}$ is the input vector. 
Thus, the loss function of Logitboost can be written out
\begin{equation*}\label{Logitboost}
L = \sum_{i=1}^{N} L_{i},      L_{i} = - \sum_{k=0}^{K-1}r_{i,k} \log p_{i,k},
\end{equation*}
where  $r_{i,k} = 0$ if $y_{i} \neq k$  and $r_{i,k} = 1$ on the contrary. A stagewise model follows as
\begin{equation*}\label{Logitboost}
F_{i,k} = F_{i,k} + v\frac{K-1}{K} (f_{i,k} - \frac{1}{K} \sum_{k=0}^{K-1} f_{i,k}),
\end{equation*}
where $v$ is a shrinkage parameter, and $f_{i,k}$ is the objective function. 

Beside the class probabilities, another important factor in Logitboost is the dense Hessian matrix, which is obtained by computing the tree split gain and node value fitting. However, certain modifications are required in order to incorporate these factors into optimization. The sum-to-zero constraint of classifier, implied by the sum-to-one constraint of the class probabilities, can be settled by adopting a vector tree at each boost. In the vector tree, a sum-to-zero vector is fitted at each split node in the K-dimensional space. Moreover, adding the vector tree allows explicit computations of the split gain and node fitting, which becomes a secondary problem when fitting a new tree. Such secondary problems can then be used to cope with the dense Hessian matrix, where only two coordinates are allowed for each of the secondary problems \citep{Sun2012}.

\subsection*{LAD Regression}
The LAD regression proposed by \cite{Friedman2002} has its loss function as $L(y,F) = |y-F|$, where $F(x)$ is solved by
\begin{equation*}\label{LAD regression}
F_{m}(x) = F_{m-1}(x) + \sum_{j-1}^{J} \gamma1 (x \in R_{jm}), \ \ \ \gamma_{jm} = \rho_{m}b_{jm}.
\end{equation*}
Moreover, in the LAD regression, the gamma parameter is
\begin{equation*}\label{LAD Optimization}
\gamma_{jm} = \median_{x_{i} \in R_{jm}} \{y_{i} - F_{m-1}(x_{i})\}.
\end{equation*}

\subsection*{M-Regression}
M-Regresison \citep{Friedman2002} is designed to incorporate with the Huber loss function
\begin{equation*}\label{M-Regression}
\gamma_{jm} = \widetilde{r_{jm}} + \frac{1}{N_{jm}} \sum_{x_{i} {\in R}_{jm}} \sign(r_{m-1}(x_{i}) - \widetilde{r_{jm}}) \bullet \min(\delta_{m},\abs(r_{m-1}(x_{i}) - \widetilde{r_{jm}})),
\end{equation*}
where $\widetilde{r_{jm}} = \median_{x_{i} \in R_{jm}} \{r_{m-1}(x_{i})\}$ and $r_{m-1}(x_{i}) = y_{i} - F_{m-1}(x_{i}).$

\subsection*{Two Class Logistic Regression}
The loss function applied in the logistic regression is a binary function \citep{Friedman2002}, which is the Bernoulli loss function. Approximately, the line search of the logistic regression can be solved from Bernoulli loss
\begin{equation*}\label{Logistic Regression}
\gamma_{jm} = \sum_{x_{i} \in R_{jm}} \widetilde{y_{i}} \bigg/ \sum_{x_{i} \in R_{jm}} |\widetilde{y_{i}}| (2-|\widetilde{y_{i}}|), j=1, \ldots, J,
\end{equation*}
where $\widetilde{y_{i}} = 2\widetilde{y_{i}} / (1+\exp(2y_{i}F_{m-1}(x_{i}))), i =1, \ldots, N.$  

\subsection*{Multiclass Logistic Regression}
The loss function applied in the Multi-class logistic regression are as follows
\begin{equation*}\label{Mutliclass logistic loss}
L(\{y_{k},F_{k}(x)\}_{1}^{K})  = - \sum_{k=1}^{K}y_{k}\log p_{k}(x), \ \ \ y_{k} = 1 \in \{0,1\}, \ \ \ p_{k}(x) = Pr(y_{k} = 1 | x),
\end{equation*}
where the line search of the multi-class logistic regression is
\begin{equation*}\label{Multiclass logistic line search}
\gamma_{jkm} = \frac{K-1}{K} \frac{\sum_{x_{i} \in R_{jkm}} \widetilde{y_{ik}}}{\sum_{x_{i} \in R_{jkm}} |\widetilde{y_{ik}}| (1-|\widetilde{y_{ik}}|)}.
\end{equation*}

\section{Ranking Problem}
One of the most discussed problems in machine learning is teaching a computer to rank. Two sets of data are required before constructing a ranking algorithm \citep{Zheng2008}, i.e., the preference data containing a set of features, and the ranked targets. Based on these two datasets, the ranking function can be computed for each dataset under an optimization problem. 

The objective function for the ranking problem is
\begin{equation*}\label{Empirical Risk}
R(h) = \frac{w}{2} \sum_{i=1}^{N} (\max\{0,h(y_{i}) - h(x_{i}) + \tau \})^2 + \frac{1-w}{2} \sum_{i=1}^{n} (l_{i} - h(z_{i}))^2,
\end{equation*}
where $x_{i}$ and $y_{i}$ are the features in the preference data, and $h(x_{i}) \leq h(y_{i}) + \tau,$ if $x_{i}$ is ranked higher than $y_{i}$.

\cite{Wu2008} proposed a highly effective ranking algorithm LambdaMART which integrates LambdaRank function and boosting. The LambdaRank function aims to maximize the Normalized Discounted Cumulative Gain (NDCG)
\begin{equation*}\label{NDCG}
N_{i} \coloneqq n_{i}\sum_{j=1}^{T} (2^{r(j)} -1) / \log(1+j),
\end{equation*}
where $r(j)$ represents the ranking of the targets. 
Gamma gradient is used in the optimization
\begin{equation*}\label{Gamma Gradient}
\gamma_{i,j} \coloneqq S_{ij} \bigg|\triangle NDCG \frac{\delta C_{ij}}{\delta o_{ij}}\bigg|,
\end{equation*}
where $S_{ij}$ takes the value of 1 or -1 depending on the relevance of the items. 
For example, in ranking for webpages, the gamma gradient is used to determine the relevance of information retrieved online. If a piece of information $i$ is more relevant than another piece $j$, then $S_{ij}$ equals to 1; otherwise $S_{ij}$ equals to -1. The $o_{ij}$ represents the difference between the ranking scores predicted by the ranking function $o_{ij} \coloneqq F(x_{i}) - F(x_{j}),$ and $C_{ij} \coloneqq C(o_{ij}) = F(x_{j}) - F(x_{i}) + \log(1+e^{s_{i} - s_{j}}).$ Moreover, the gamma gradient of a specific item $i$ is as follows
\begin{equation*}\label{Gamma Gradient Summation}
\gamma_{i} = \sum_{j \in P} \gamma_{ij}.
\end{equation*}

\section{Conclusion}
In this paper, we summarize the Gradient Boosting Algorithms from several aspects, including the general function optimization, the objective functions, and different loss functions. Additionally, we present a set of boosting algorithms with unique loss functions, and we solve their predictive models accordingly.

\end{document}